RESEARCH ARTICLE OPEN ACCESS

# Embarrassed to Observe: The Effects of Directive Language in Brand Conversation

Andria Andriuzzi[1] | Géraldine Michel[2]

[1]Coactis, Université Jean Monnet, Saint-Etienne, France | [2]IAE Paris-Sorbonne, Université Paris 1 Panthéon-Sorbonne, Paris, France

**Correspondence:** Géraldine Michel (geraldine.michel@iae.pantheonsorbonne.fr)

**Received:** 30 July 2024 | **Revised:** 15 July 2025 | **Accepted:** 17 July 2025

**Funding:** The first author received funding from Saint-Etienne Métropole. The funding sources had no role in the conduct of the research.

## ABSTRACT

In social media, marketers attempt to influence consumers by using directive language, that is, expressions designed to get consumers to take action. While the literature has shown that directive messages in advertising have mixed results for recipients, we know little about the effects of directive brand language on consumers who see brands interacting with other consumers in social media conversations. On the basis of a field study and three online experiments, this study shows that directive language in brand conversation has a detrimental downstream effect on engagement of consumers who observe such exchanges. Specifically, in line with Goffman's facework theory, because a brand that encourages consumers to react could be perceived as face-threatening, consumers who see a brand interacting with others in a directive way may feel vicarious embarrassment and engage less (compared with a conversation without directive language). In addition, we find that when the conversation is nonproduct-centered (vs. product-centered), consumers expect more freedom, as in mundane conversations, even for others; therefore, directive language has a stronger negative effect. However, in this context, the strength of the brand relationship mitigates this effect. Thus, this study contributes to the literature on directive language and brand–consumer interactions by highlighting the importance of context in interactive communication, with direct relevance for social media and brand management.

## 1 | Introduction

Social media are critical for brands as they are key platforms where users, including brands and consumers, can create and share content (Kaplan and Haenlein 2010). On social networking sites such as Facebook, X, and Instagram, text-based interaction through user comments is ubiquitous. In the United States, 45% of social media users "like" brand posts, and 22% share brand-related content (Sprout Social 2021). Similarly, shares and comments are manifestations of consumers' engagement with brands (Gavilanes et al. 2018). These engagement metrics are tracked by 78% of marketers (Sprout Social 2023), which reveals their importance for marketing practice.

The most prominent form of engagement occurs when consumers comment on brand posts and when brands answer consumers' posts, creating a form of conversation. Brand conversations are unscripted verbal interactions between brands and consumers on social media (Andriuzzi and Michel 2021). Such conversations can be critical for a brand's image and reputation: 51% of U.S. consumers believe that the "most memorable" brands are those who respond to customers via social media (Sprout Social 2023), even on broader subjects such as current events, social issues, industry trends, and lifestyles (Arrivé 2022).

When interacting with consumers, brands frequently use directive messages (Villarroel Ordenes et al. 2019), a popular technique for encouraging social media users to act, or to react to conversations (e.g., on X—formerly Twitter: "Use the BK app today"—Burger King; "Finish either of these sentences and you could win!"—Dunkin Donuts). Despite the wide use of directive messages in brand conversations (as opposed to single-message strategies, like those used in advertising), we know little about their impact on consumers who, at a specific time, see brands interacting with other consumers in such conversations. Nevertheless, this is important because, on social media, consumers generally consume more content than they interact with (Campbell et al. 2014). As an example, while 42% of U.S. consumers said they followed a brand on social media in exchange for an incentive (e.g., a discount code), only 26% reported commenting on a brand's post (Marketing Charts 2022). Even in the context of less instrumental relationships, such as in online brand communities, "lurkers" represent a greater proportion (55%) than "posters" do (45%; Mousavi and Roper 2023). However, "it is important to support lurkers as well, because there is no significant difference between posters and lurkers in the outcomes of purchase intention, positive WOM [word-of-mouth], and (…) resistance to negative information" (Mousavi and Roper 2023, 82). Moreover, viewing online brand content can lead consumers to engage (Ellison et al. 2020). Thus, in the expanding social media brand landscape, in which interactivity plays a large role, it is relatively surprising that we still know little about the impact of different brand interaction styles on consumers who see such brand–consumer interaction on their social media feeds.

The marketing literature focuses on the use of directive language—that is, the use of expressions designed with the goal of provoking a reaction from the recipient (Searle 1979)—in the context of online advertising. Termed "calls to action" (CTAs) by practitioners, directive messages are supposed to increase web traffic and conversion. CTAs are widely used by marketers (Semrush 2022), and research shows the positive impact of this powerful language on engagement metrics (Pezzuti et al. 2021). However, research in advertising also shows mixed results from the use of directive language, for example, depending on the brand relationship, product, and consumer characteristics. In fact, for committed consumers, assertive ads create greater pressure to comply, which reduces compliance and ultimately leads to decreasing ad and brand liking (Zemack-Rugar et al. 2017). In another vein, for high-power consumers, assertive ads are effective in promoting "want" products but ineffective in promoting "should" products. On the contrary, for low-power consumers, assertive ads are effective in promoting "should" products but ineffective in promoting "want" products (Wang and Zhang 2020).

Even if the research mentioned above explains consumer reactions to directive language in the context of advertising, we know little about the (in)effectiveness of addressing consumers this way in brand conversations. Managerial literature is also inconclusive. However, in community management, it is especially important to know which language form should be used when addressing consumers (Carnevale et al. 2017; Packard et al. 2024). As encouraging and shaping consumer participation are crucial for brands' market success, or even for social change (Fletcher-Brown et al. 2024; Vredenburg et al. 2020), we wonder what is the impact of directive language on consumers who see brands addressing other consumers this way in brand conversations.

To explore the impact of directive brand language, we need to consider the different types of brand conversation that involve different linguistic norms (Jakic et al. 2017). Online brand–consumer conversations can be classified into two main categories: product-centered (conversations that typically feature informative content directly related to the brand products, Lee et al. 2018) and nonproduct-centered (content related to the brand's universe that does not directly refer to its products, Arrivé 2022). The latter type is a common practice for brands on social media. For instance, Dove celebrates the diversity of women's bodies and beauty rather than just mentioning their personal care products. Similarly, Nike shares inspiring stories about athletes and behind-the-scenes content that highlights resilience and diversity beyond the showcase of its latest sneaker (Doan 2023). Research suggests that, in product-centered conversations, a rather direct communication style is needed (Köhler et al. 2011). This raises the question of whether the impact of directive language in brand conversations that focus on products/services (product-centered) differs from that concerning broader topics (nonproduct-centered).

Finally, we also need to consider brand relationship strength. Research suggests a deep interplay between consumer–brand interactions and brand relationships (Hudson et al. 2016). Brand relationship norms influence consumer behavior, as such individuals are more likely to interact directly with brands if they have an emotional relationship with them (Aggarwal 2004). Consequently, this study also aims to better understand the role of brand relationship strength in the effects of directive brand language.

To answer these questions, we use facework theory (Goffman 1967), which emphasizes that, in mundane conversations, directive expressions such as orders, requests, and advice can be assimilated to "face threats" and lead to negative outcomes (Brown and Levinson 1987). We use facework theory because it allows us to gain a deep understanding of how consumers behave when they witness the violation of conversational norms, even without necessarily participating in the conversation. Consequently, in light of facework theory, the goal of this study is to better understand how consumers respond when they see a brand using directive language in the context of brand–consumer conversations on social media.

The present research relies on one field study (271,345 posts), and 3 experiments ($N = 677$). On the basis of seven major fast food brands' activities on X, our field study reveals an overall negative effect of directive brand language on consumer engagement behavior (e.g., "likes"). The first two experiments highlight the role of vicarious embarrassment in explaining the negative effect of directive language on consumer engagement. Indeed, when consumers observe the use of inappropriate directive language by brands towards others, they can feel a vicarious embarrassment about a situation that is undesirable for others (Miller 1987). We also find that directive brand language has a more detrimental effect on consumer engagement when the conversation is nonproduct-centered than when it is product-centered. Finally, a third experiment highlights the moderating role of brand relationships, helping to better understand the ambivalent effects of directive brand language use on consumers. Therefore, this study contributes to the literature on directive language, brand–consumer interactions and engagement, and facework, with managerial relevancy for practitioners such as brand, social media, and community managers.

## 2 | Theoretical Background

### 2.1 | Directive Language in Brand Conversation

According to speech act theory, language is used primarily to achieve actions (Searle 1969). The expressions used by individuals have three dimensions: a locutionary function (the literal meaning), an illocutionary function (the speaker's intention in saying something), and a perlocutionary function (the actual effect of the utterance on the recipient; Austin 1962). Building on Austin, Searle (1979) defines five types of illocutionary speech acts that are directly related to perlocutionary effects: (1) *assertives* are statements that convey information or describe something, such as a brand claiming, "Our products reduce energy consumption by 20%," with the goal of influencing purchase decisions; (2) *directives* are intended to motivate the listener to do something, such as requests or commands, that is, asking consumers to "Sign up for our newsletter for exclusive discounts!" to encourage immediate action; (3) *commissives* are when the speaker commits to doing something in the future, such as a brand promising that "We will reduce our carbon footprint by 50% in the next 5 years," which could help build trust; (4) *expressives* convey the speaker's emotions or feelings, such as a brand saying, "We are thrilled to have such a great community!" to build loyalty; finally, (5) *declarations* actually change reality by being spoken, such as announcing a corporate merger or a product launch, informing stakeholders of significant changes.

In digital communication, marketers frequently use directive messages with the aim of influencing consumer behavior, as in CTAs (Vafainia et al. 2019). Directive brand messages typically consist of orders containing imperative verbs and exclamation marks. They can also consist of more nuanced suggestions, for example, using the verb "would" (Brown and Levinson 1987), or asking a question instead of making a command. Indeed, "questions are a subclass of directives, since they are attempts by [the speaker] to get [the hearer] to answer, that is, to perform a speech act" (Searle 1979, 14).

Directive messages in the form of CTAs are recommended in the managerial literature (Crowdfire 2022; Hootsuite 2022; Hubspot 2024). However, previous academic research has shown mixed results. On the one hand, directive messages can have a negative effect on the likes received by brands depending on the brand–consumer relationship: brand-committed consumers can feel some pressure to comply with directive brand language and therefore react negatively, as suggested by reactance theory (Zemack-Rugar et al. 2017). On the other hand, the impact of directive ads can depend on the product type and consumer characteristics. For example, directive ads for hedonic products can lead to higher compliance among consumers with high power beliefs; conversely, directive ads for nonhedonic products can lead to higher compliance among low-power consumers (Wang and Zhang 2020). Research has shown that emotional or informative posts lead to more consumer sharing than directive posts do (Villarroel Ordenes et al. 2019). Other studies have revealed that asking for likes and comments increases both of those metrics, whereas asking questions increases comments but decreases likes (Lee et al. 2018). Finally, research shows that, in direct marketing (i.e., emailing), CTAs have more positive effects when they feature nonmonetary incentives (vs. monetary incentives; Vafainia et al. 2019).

Despite a decent body of research on directive language in advertising, research examining directive language in the context of brand–consumer conversations is scarce. Indeed, most of the research mentioned above focuses on single-message strategies, without considering the interactive dimension of digital communication and the presence of other consumers (Table 1). However, consumer participation heavily influences the subsequent consumer behavior (Ben and Shukla 2024; Hamilton et al. 2017). Thus, we wonder how consumers react when they observe a brand using directive language to address other consumers in social media conversations.

Among the reactions that are desired by marketers, engagement is a concept presented as the propensity for consumers to positively interact with a brand (Baldus et al. 2015; Dessart et al. 2015) as well as the psychological state experienced during such an interaction (Hollebeek et al. 2014). Consumer engagement is an intentional allocation of resources, such as cognitive, emotional, and behavioral resources by the consumer during their interaction with a particular object provided by an organization (Elmashhara et al. 2024; Hollebeek et al. 2022).

While the broad concept of engagement has three dimensions (as stated: emotional, cognitive, and behavioral; Dessart et al. 2016), this study focuses on the impact on behavioral engagement. In doing so, we draw on recent conceptual developments of engagement, which state that one of its pillars lies in "content engagement" (Obilo et al. 2021), a concept relevant to consumer responses to brand posts on social media. Indeed, in digital communications, the behavioral dimensions of consumer engagement include sharing, "liking," or commenting on brand content (Swani and Labrecque 2020). Research shows that such engagement can be determined by the characteristics of the brand post (De Vries et al. 2012) and by a more or less proactive attitude displayed by the brand (Schamari and Schaefers 2015). On social media, brands often explicitly call for consumer participation by using directive language in their posts and answers to consumers (e.g., "Give me your opinion!", "Tell us which color you prefer!"). However, it remains unclear whether such strategies actually help drive more engagement behaviors, especially when brands use them in a conversational context.

### 2.2 | Facework and Vicarious Embarrassment

To address our research questions, we build on Goffman (1967) facework theory. According to facework, when individuals

**TABLE 1** | How the current research complements existing research on directive brand language.

| Research | Object | Moderating variables | | | Dependent variable | Main results |
| --- | --- | --- | --- | --- | --- | --- |
| | Conversation (two-way interaction) | Conversation topic | Brand relationships | | Consumer engagement | |
| This study | Yes | Yes | Yes | | Yes | Directive language causes vicarious embarrassment and reduces engagement in nonproduct brand conversations. Brand relationships mitigate this effect. |
| De Vries et al. (2012) | No | No | No | | Yes | Call to act, defined as "medium-level interactive brand post," has no significant effect on likes or comments. |
| Irmak et al. (2020) | No | Yes | No | | No | Government-level messages have a negative effect on conservative consumers except in the case of a notification (vs. a warning). |
| Kavvouris et al. (2020) | No | No | No | | No | Injunctive (vs. descriptive) messages are more threatening and generate lower behavioral intentions. |
| Kronrod et al. (2012) | No | Yes | No | | No | Directive messages lead to more compliance when the cause seems to be important. |
| Lee et al. (2018) | No | Yes | No | | Yes | Asking for likes and comments has a positive effect on both metrics. Questions increase comments but decrease likes. |
| Meire et al. (2022) | Yes | Yes | No | | Yes | Aligning the language style of a brand's posts with that of its community improves engagement, especially for unfamiliar content. |
| Oltra et al. (2022) | No | No | Yes | | No | Calls to action can increase consumer inspiration, especially among those with low brand identification. |
| Pezzuti et al. (2021) | No | No | Yes | | Yes | Consumers engage more when brands use certainty words because they feel like the brand has more power. |
| Pogacar et al. (2018) | No | No | No | | No | Congruency between marketing and message elements helps to understand and enhance the effectiveness of linguistic devices. |
| Razzaq et al. (2023) | No | No | No | | Yes | In meme marketing, combining directive speech acts with emotional images and assertive messages can reduce perceived dominance. |

(Continues)

TABLE 1 | (Continued)

| Research | Object | Moderating variables | | | Dependent variable | Main results |
| --- | --- | --- | --- | --- | --- | --- |
| | Conversation (two-way interaction) | Conversation topic | Brand relationships | | Consumer engagement | |
| Septianto and Garg (2021) | No | No | No | | No | Inducing gratitude increases compliance with directive messages (e.g., promoting responsible drinking) by reducing psychological reactance. |
| Vafainia et al. (2019) | No | No | Yes | | No | Nonmonetary incentives in direct mail have more impact than monetary ones. Customer heterogeneity increases the effect of CTAs. |
| Villarroel Ordenes et al. (2019) | No | No | No | | Yes | Emotional and informative brand messages are more shared than directive ones. |
| Zemack-Rugar et al. (2017) | No | No | Yes | | No | Consumers with strong brand relationships react negatively to ads that instruct them to take action, because it creates pressure to comply. |

interact, they try to maintain their own face and, at the same time, ensure that the other participants do not lose face (Goffman 1967)—where face is defined as "the positive social value a person effectively claims for himself by the line others assume he has taken during a particular contact" (Goffman 1955, 213). To achieve these goals, they use several strategies aimed at avoiding or minimizing face-threatening acts (FTAs; Brown and Levinson 1987). FTAs typically include imperative talk, such as orders, requests, and even suggestions or advice. Facework theory highlights that, in mundane conversations, such directive language can lead to a reduction in people's freedom and autonomy, a situation that causes reactance (Brehm 1966) and leads to negative outcomes in subsequent interactions (Brown and Levinson 1987).

We use facework theory because, first, it is particularly relevant in the context of consumer–brand conversations. Other competitive theories, such as speech acts (Austin 1962; Searle 1969), also study the impact of linguistic devices such as orders or requests and their effects on recipients. However, "Austin's and Searle's approaches have in common that they concentrate on single communicative acts, [with] as a certain danger inherent in this, namely, that it is easy to lose perspective on communication as a whole. After all, communicative acts seldom occur in isolation, but rather sequentially in interaction" (Allwood 1977, 10). Conversely, Goffman (1967) facework applies to whole interaction sequences during social encounters instead of being single-message focused. Moreover, it also applies to the online context (Park 2008). Because consumer–brand conversations resemble human-to-human conversations, using the same technical (i.e., social media platforms) and linguistic (i.e., human language) devices, facework is relevant to the study of online brand–consumer conversations instead of the study of single-message brand strategies.

Second, facework is particularly relevant to the study of consumers who are not the direct recipients of directive language but see such brand–consumer conversations instead, even if not participating in it. The literature frequently uses reactance theory to explain that recipients of directive messages can feel that their freedom is reduced and, thus, avoid the recommended option (Brehm 1966; Zemack-Rugar et al. 2017). Even if facework shares close concerns with reactance theory, reactance may be less relevant to explain consumers' responses when they see a brand using directive language in a conversation with another consumer. Indeed, consumers may be unlikely to feel that their freedom is limited by a message that is not directly addressed to them. Nevertheless, facework theory provides an explanation for the potential negative reactions of consumers toward directive language addressed to others: according to facework, when consumers see other people being addressed in ways that do not conform to conversational norms (e.g., by being the target of FTAs such as orders), they may feel *embarrassed* by the loss of face of the recipients of directive messages (Goffman 1956). As Goffman (1956, 265) states: "When an individual finds himself in a situation that ought to make him blush, others present will usually blush with and for him."

This embarrassment for others is called vicarious embarrassment, that is, an uncomfortable state felt when witnessing a person in an undesirable situation (Miller 1987). Consumers can feel vicarious embarrassment in sensitive purchase contexts (Ziegler et al. 2022) or when witnessing the violation of social norms in both customer-to-customer and customer-to-employee interactions (Kilian et al. 2018). In terms of the empathy felt by consumers, vicarious embarrassment is different from disappointment with a brand, which is related to a feeling of deception regarding some expectations (Tan et al. 2021).

Vicarious embarrassment felt by consumers can lead to negative outcomes, such as negative word-of-mouth and a negative impact on the overall image of the service provider (Kilian et al. 2018; Ziegler et al. 2022). It can also cause consumers to disengage from the brand's social media when they experience socially unacceptable behavior (Villanova and Matherly 2024). Thus, when brands infringe on the social norms of conversation by addressing others with directive language, consumers may feel vicarious embarrassment due to the loss of face of the recipients. People can experience vicarious embarrassment purely due to the way language is used in an interaction, even if the topic itself is not inherently sensitive. This vicarious embarrassment often arises from violations of conversational norms, unexpected directness, or socially awkward phrasing, leading to discomfort independent of topic valence (Goffman 1956). As a result, consumers may avoid engaging in situations that feel uncomfortable and may even want to repair the social image or face of others (Goffman 1956), especially when such a situation occurs in public (Sangwan et al. 2024). By avoiding engaging with the brand, consumers distance themselves from the negative situation and signal their disapproval of the brand's actions.

We focus here on consumers' behavioral engagement (Hollebeek et al. 2022), that is, consumers' reactions to brand conversations on social media, such as subsequent interactions in the conversation, sharing brand posts, or liking brand posts. Considering the above development, we assume that directive language use could induce vicarious embarrassment leading consumers to reduce their behavioral engagement in a discussion in which a brand is seen as an agent addressing other consumers in an undesirable way. Consequently, we formally posit the following hypotheses:

**H1.** *Directive brand language addressed to others in a brand conversation has a detrimental effect on consumer engagement.*

**H2.** *Vicarious embarrassment mediates the negative effect of directive brand language on consumer engagement.*

## 2.3 | Product- Versus Nonproduct-Centered Conversation

To explore the impact of directive brand language, we need to consider the different types of brand conversation that involve different linguistic norms (Jakic et al. 2017). Conversation has been shown to be an antecedent to online brand community engagement, particularly in terms of two-way interaction (Zhao and Chen 2022). This is especially important today, as brands initiate conversations connected to mundane topics or to social/environmental causes, which require specific language (Fletcher-Brown et al. 2024).

Online brand–consumer conversations are often classified into two main categories: product-centered conversations and nonproduct-centered conversations. This classification aligns with the associations that consumers make about product attributes: product-related associations (e.g., information about product ingredients) and nonproduct-related associations (i.e., "external aspects of the product or service that relate to its purchase or consumption," Keller 1993, 4). On the one hand, product-centered conversations typically feature directly informative content about the product, such as "details about products, promotions, availability, price, and product-related aspects that could be used in optimizing the purchase decision" (Lee et al. 2018, 5107). On the other hand, nonproduct-centered conversations consist of "casual banter, or discussion of the brand's philanthropic outreach" (Lee et al. 2018), or other content related to the brand's universe not directly referring to its products (Arrivé 2022).

Research suggests that, in product-centered conversations, a rather direct communication style is needed. For example, when interacting with a virtual agent about financial services, new users prefer a conversation that features functional content from a proactive agent (Köhler et al. 2011), which likely involves the use of clear, direct instructions. However, other consumers seek interactions stemming from social motivations rather than functional motivations (Ashley and Tuten 2015). Thus, consumers' expectations in nonproduct-centered conversations should differ from those in product-centered conversations. In such a context, the brand conversation resembles a mundane consumer-to-consumer conversation; thus, facework theory suggests that directive language would likely damage the continuation of ongoing conversation by damaging the autonomy of the consumer (Brown and Levinson 1987). Thus, nondirective language would be here the expected norm.

In product-centered conversation, consumers often tolerate or even appreciate directive language because such language can be perceived as informational or improvement-oriented. Thus, refraining from directive language may foster more positive responses in nonproduct-related interactions than in product-related conversations. In other words, consumers may react more negatively to directive language in a nonproduct-centered conversation—a context in which such language is not expected—than in a product-centered conversation. Therefore, we formally posit the following:

**H3.** *Directive brand language drives less consumer engagement when the conversation is nonproduct-centered than when it is product-centered.*

## 2.4 | Brand Relationship Strength

According to facework theory, individuals consider social distance when choosing their interaction strategies (Brown and Levinson 1987). For example, when interacting with close relatives, the risk of losing face is low; therefore, a direct, "bald-on-record" language is appropriate. Conversely, when interacting with strangers, the risk of losing face is high; thus, individuals may avoid directive language that could damage the hearer's face.

As some consumers can feel that they have intimate relationships with their favorite brands (Fournier 1998), facework theory suggests that the impact of directive language depends on the consumer's relationship with the brand. Like with close relatives, direct brand language would even be the expected norm for consumers who have a close relationship with a brand. Such committed consumers may feel like they are part of a community and therefore feel connected not only with the brand but also with other consumers who interact with the brand's social media content (Dessart et al. 2015). Because of this communal connection, consumers may identify with interactants and therefore agree with the use of direct language by the brand, as it is the expected norm in this situation.

In addition to praising the brand for respecting conversational norms, consumers engaged in close brand relationships could also show tolerance when the brand deviates from norms. Indeed, brand relationship quality influences consumer responses not only in direct interactions but also when consumers observe interactions with others. Fournier (1998) shows that strong emotional ties lead consumers to adapt to and tolerate brand transgressions. Thus, close brand relationships increase forgiveness for brand missteps (Donovan et al. 2012), promoting tolerance from consumers who are more likely to justify the brand's actions, even those they have just witnessed.

A strong type of brand relationship, brand love, also leads to loyalty and resistance to negative information (Batra et al. 2012), forgiveness of mistakes, and positive word-of-mouth (Rahman et al. 2021). Therefore, consumers in a strong brand relationship are more likely to forgive a brand's faux pas in conversations with other consumers and may even want to continue to engage. Therefore, we posit the following:

**H4.** *Brand relationship strength mitigates the negative effect of directive brand language on consumer engagement.*

Figure 1 summarizes our conceptual framework.

## 3 | Overview of Studies

The first study uses X web scraping to address the basic premises of this study, where we posit a negative effect of directive language on consumer engagement (H1; Table 2). In addition, its open API facilitates large-scale textual analysis (at the time of data collection), as data availability is an issue for such methods (Berger et al. 2020). However, since the next studies employ experiments instead of web scraping, we were able to study directive language on other platforms where comments are also important, that is, Facebook and Instagram. These other contexts add to the external validity of the research.

The second study aims to explain the proposed main effect so that when consumers see a brand using directive language directed at other consumers on social media, they feel vicarious embarrassment, which has a detrimental effect on their engagement with the brand (H2). We also test the idea that when a brand engages in a conversation that is nonproduct-centered, directive language triggers less engagement from consumers (H3). Finally, the third study offers a complementary understanding of the relationship between directive language use and consumer engagement by examining the moderating role of brand relationship strength (H4).

Our studies focus on a single industry, that is, food, to control for specific patterns and dynamics of social media engagement that may vary across product categories (De Vries et al. 2012). We chose the fast food industry for studies 1 and 2 because consumers have been shown to voice their opinions about fast food brands on social media (Šerić and Praničević 2018). This situation leads practitioners to carefully manage consumer comments to maintain their reputations (Kim and Velthuis 2021). Responding to consumers generally leads to more engagement (Schamari and

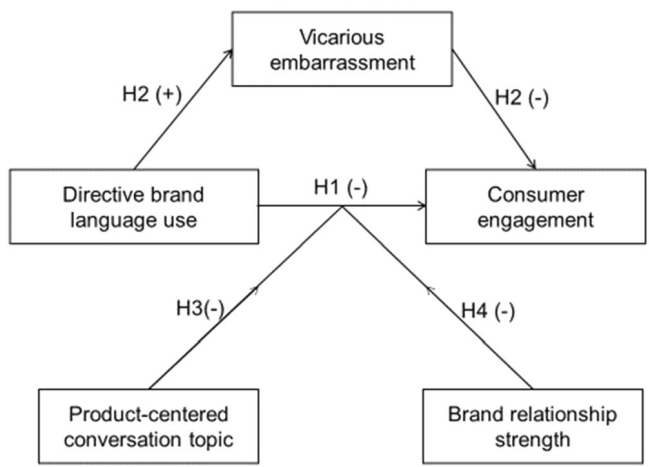

**FIGURE 1** | Conceptual framework.

**TABLE 2** | Study overview.

| Study | Procedure | Sample | Industry | Brand | Hypotheses |
|---|---|---|---|---|---|
| S1 | Field study (X, ex-Twitter) | 271,345 | Fast-food services | 7 major brands | H1 |
| S2a | 2 × 2 online experiment (consumer panel; mock Facebook post and comments) | 281 | Fast-food services | McDonald's | |
| S2b | 2 × 2 online experiment (Prolific; mock Facebook post and comments) | 275 | Fast-food services | McDonald's | H1, H2, H3 |
| S3 | 2 × 2 online experiment (consumer panel; mock Instagram post and comments) | 121 | Food products | 5 major brands (participant-chosen) | H1, H4 |

Schaefers 2015), and this two-way interaction creates a form of brand conversation that is central to our research. While first examining food services in studies 1 and 2, we switched to food products for study 3 to increase external validity while remaining in a related industry for consistency.

The choice of the food category was also critical to the success of the study designs. First, for the field study, sufficient data were needed to conduct our text mining and analysis (Berger et al. 2020). Sufficient amounts of data can be found in fast food categories where, as mentioned earlier, well-known brands and their consumers are quite active. Second, it was critical for our experiential study designs to create realistic scenarios to enhance the experience for participants and increase the external validity of the studies (Aguinis and Bradley 2014). Therefore, scenarios in which a brand and a consumer interact in a category where both consumers and brands are active on social media provide realistic ground.

# 4 | Study 1: Field Study on the Detrimental Effect of Directive Language on Consumer Engagement

The first study uses a field study environment to detect the use of directive language within brand reply posts and determine whether such messages have an impact on consumer engagement. To do so, we used a large data set of real brand replies and relied partly on natural language processing (NLP; Berger et al. 2020).

Brand participation in online conversations plays a crucial role in shaping consumer perceptions (Homburg et al. 2015) and behaviors. Overall, the more a brand interacts with consumers, the stronger the positive effects on sales and customer relationships (Kumar et al. 2016). Brand responsiveness also encourages consumer participation by portraying the brand as open and willing to engage (Smith et al. 2012). Specifically, brand responses to online complaints improve brand evaluations (Van Noort and Willemsen 2012) and lead to positive consumer behaviors, who are more inclined to continue buying and recommending the brand (Kim et al. 2016). Similarly, responses to positive comments enhance engagement from consumers (Schamari and Schaefers 2015). However, the way in which brands respond—rather than simply whether they respond—has a significant effect on the way in which consumers engage. For example, in a study of negative customer reviews on TripAdvisor, personalized responses from hotel managers had a positive effect on the valence of subsequent reviews (Wang and Chaudhry 2018). Therefore, the language and style used by brands in their responses must be carefully considered.

By focusing on brand responses to consumers on X, we aim to understand how these interactions affect engagement from other users. When an X user sees a brand responding to another consumer's post, to what extent would they be inclined to engage, even if they are not the recipient of the response? Focusing on engagement with brand responses to posts—not posts initially addressed by brands to all users—allows us to clearly differentiate the object of our research (the effect of brand language in conversations with consumers) from general brand messages (e.g., promotional posts, contests, and brand news). Moreover, the visible engagement metrics (likes, comments, shares) available on X enable us to measure the engagement of consumers with a brand's response to another consumer.

## 4.1 | Data

We collected all posts from seven major brands in the fast food industry ($N = 273{,}827$; Table 3) that were produced over a period of 2 years (April 1, 2021, through March 31, 2023). All seven accounts are the main accounts of U.S. brands and employ English to interact with consumers. The targets are thus U.S. consumers, as these brands have other accounts and use the local language when operating in other countries.

To collect posts, we used the Python module "nscrape," which is specifically designed to extract data from social media sites. Our extracted data included each post's content with related information such as the publication time and date. In addition, we performed a sentiment analysis[1] of brand posts to determine if the reply posts were negative, neutral, or positive.

Since our main interest is in how consumers react when they see a brand using directive language when addressing another consumer, we focused on brand posts that were direct responses to specific users (i.e., "@reply" posts, as opposed to regular posts addressed to all brand followers) identified by the "@" symbol used at the beginning of the post, immediately followed by a user's name. This left us with a "@reply" post data set of 271,345 posts.

TABLE 3 | Field study sample.

| Brand | All posts | | Regular posts | | @Reply posts | |
|---|---|---|---|---|---|---|
| Burger King | 17,873 | 6.5% | 364 | 14.7% | 17,509 | 6.5% |
| Dunkin' | 8136 | 3.0% | 354 | 14.3% | 7782 | 2.9% |
| KFC | 3230 | 1.2% | 278 | 11.2% | 2952 | 1.1% |
| McDonald's | 151,142 | 55.2% | 260 | 10.5% | 150,882 | 55.6% |
| Pizza Hut | 59,865 | 21.9% | 400 | 16.1% | 59,465 | 21.9% |
| Starbucks | 24,882 | 9.1% | 292 | 11.8% | 24,590 | 9.1% |
| Subway | 8699 | 3.2% | 534 | 21.5% | 8165 | 3.0% |
| Total | 273,827 | 100% | 2482 | 100% | 271,345 | 100% |

As a proxy for the use of directive language, we focused on the use of the imperative mood in social media brand reply posts. Indeed, in natural language, the imperative is used with the aim of inducing the receiver to act in a way that is intended by the speaker. Using the imperative form is "the most direct approach" when addressing someone with the aim of expressing a need (Yule 1996, 63). To detect the presence of imperatives in brand reply posts, we conducted a natural language processing study using the STANZA library (Qi et al. 2020), a Python model for automated natural language processing. This model is trained for morphosyntactic annotation tasks via a corpus of three million words. It can be used to perform a variety of linguistic analysis tasks, including verb detection, tense, and mood (in our case, imperative). Thus, our analysis allowed us to categorize each word in each brand reply post to determine whether it was a verb and, if so, whether it was in the imperative form. Each reply post was annotated according to the presence or absence of an imperative verb, using binary notation (0 for no imperative, 1 for imperative).

## 4.2 | Results

We found that 54.1% of the brand reply posts used an imperative verb ($N_{directive} = 146.747$), whereas 45.9% used no imperative verb ($N_{control} = 124.598$). We tested the relationship between the use of directive language (i.e., the presence or absence of a verb in the imperative mood) and consumer engagement through multivariate analysis of variance (MANOVA). The independent variable was a binary variable reflecting the absence or presence of an imperative verb, whereas the dependent variable, consumer engagement, was operationalized via likes, comments, reposts, and quotes.[2]

We found a significant and negative relationship between the presence of an imperative verb in brand replies on all the engagement variables: like ($F(1, 271343) = 111.61$, $p < 0.001$), repost ($F(1, 271343) = 52.79$, $p < 0.001$), reply ($F(1, 271343) = 153.91$, $p < 0.001$), and quote ($F(1, 271343) = 45.00$, $p < 0.001$). All consumer engagement means were lower when brands used the imperative mood in their reply posts than when they did not (Table 4). These results support H1. Alternatively, we ran the same analysis with brand name, brand reply post sentiment (negative, neutral, or positive), and publication date and time (year, month, day, and hour) as covariates, and the results were still consistent and significant.

## 4.3 | Discussion

This study finds support for H1. When brands address consumers using directive language within their reply posts (i.e., using the imperative mood), consumers engage less than when the posts do not include an imperative verb. Indeed, the number of likes, reposts, replies, and quotes from consumers decreases when the reply posts contain verbs in the imperative form. These results align with our initial prediction in a natural field setting using real posts from major brands, supporting a main negative effect of directive language use on consumer engagement.

Although this study provides valuable insights, it has several limitations. While selecting brand posts on the basis of the @username filter alone helps to identify brand responses, it may miss nuances such as the intent behind the messages or their context. For example, a brand message such as "@username, visit our website" may be considered directive, but it may not generate much engagement simply because it does not require much response. In contrast, a more interactive or behavioral request, such as "@username, indicate the color you purchased," might result in greater engagement. Even if we controlled for the valence of the post in an alternative analysis (not reported for brevity), the actual content of the posts could affect engagement in addition to the directive nature of the language alone.

Finally, to strengthen the robustness of our findings and because this study's setting cannot fully support a causal relationship or provide any explanation for consumer behavior, our next studies will use experimental designs to further explore the relationship between directive language use and consumer engagement.

## 5 | Study 2: Mediating Role of Vicarious Embarrassment and Moderating Role of Conversation Topic

Study 2a tests our prediction on the effect of directive language on vicarious embarrassment and the role of conversation topic. Study 2b tests the causal relationship between directive language use and consumer engagement (H1), with the mediating

TABLE 4 | Field study results.

|  | No directive language | | Directive language | | | |
| --- | --- | --- | --- | --- | --- | --- |
|  | Mean | SD | Mean | SD | F | p |
| Like | 36.2 | 1225.0 | 2.1 | 153.9 | 111.6 | < 0.001 |
| Retweet | 2.3 | 109.9 | 0.2 | 28.0 | 52.8 | < 0.001 |
| Reply | 0.9 | 16.9 | 0.3 | 4.5 | 153.9 | < 0.001 |
| Quote | 0.5 | 24.3 | 0.1 | 2.5 | 45.0 | < 0.001 |

9 of 17

role of vicarious embarrassment (H2) and the moderating role of conversation topic (H3).

## 5.1 | Study 2a

### 5.1.1 | Method

To test our initial assumptions, we designed a 2 (brand language: directive language use vs. control) × 2 (conversation topic: product-centered vs. nonproduct-centered) between-subjects experiment, with vicarious embarrassment as dependent variable. Two hundred eighty-one participants ($M_{age}$ = 42.10, SD = 12.83; 50.2% male, 49.8% female) were recruited from an online consumer panel and received market rate financial compensation. All participants lived in France. They were screened as having declared that they were active Facebook users and were randomly assigned to the different experimental cells.

We asked the participants to imagine that they were browsing their Facebook newsfeed; then, we exposed them to a stimulus, that is, a fictitious conversation between McDonald's brand and a consumer (Supporting Information S1: Web Appendix A). The conversation topic was manipulated by creating a conversation about a new product from the brand (a new organic patty; topic = product-centered) and a conversation about a broader topic (a post about practicing sport; topic = nonproduct-centered). Brand use of directive language was manipulated by including directive language in a brand conversation with another consumer (e.g., "Tell us when you will come and see us!") or by not including directive language (control). Following the scenario, the participants answered four items via seven-point Likert scales indicating the extent to which they perceived the presence of directive language (e.g., *"[Brand] incites people to answer questions,"* $\alpha$ = 0.93—all measures are presented in Supporting Information S1: Web Appendix D.). Regarding the conversation topic, we asked participants if they thought *"the conversation (was) about a product sold by [Brand]: (yes/no)."* To measure vicarious embarrassment, participants read the sentence: *"If I put myself in the shoes of the internet user [Brand] is addressing..."* and answered three dichotomous items (e.g., *"I do not feel embarrassed at all/I feel very embarrassed"*; $\alpha$ = 0.94; Dahl et al. 2001).

### 5.1.2 | Results

We checked the manipulations first by conducting a one-way ANOVA. The findings revealed a significant effect of directive language use on directiveness perception ($F(1, 279)$ = 46.16, $p < 0.001$), such that directive language perception was greater in the directive condition ($M_{directive}$ = 5.52, SD = 1.18) than in the control condition ($M_{control}$ = 4.37, SD = 1.62). Second, the question concerning the conversation topic served as a manipulation check, and also checked participants' attention to the questionnaire. Indeed, we screened participants on the basis of their attention by keeping only those who remembered the conversation topic correctly. Thus, as expected, the relationship between variables (i.e., manipulated topic and perceived topic) was significant ($\chi^2 (1, N = 281)$ = 281, $p < 0.001$). Both results prove that the manipulations were successful.

Then, we conducted a two-way ANOVA with directive language use and conversation topic as independent variables, and vicarious embarrassment as the dependent variable.[3] The findings revealed a significant and positive effect of directive language use on vicarious embarrassment ($F(1, 277)$ = 4.25 $p$ = 0.040, $\eta_p^2$ = 0.015), such as the use of directive language increased vicarious embarrassment ($M_{directive}$ = 2.61, SE = 0.13) compared with the control condition ($M_{control}$ = 2.22, SE = 0.13). The direct effect of conversation topic ($p$ = 0.930) and the interaction effect of directive language use and conversation topic ($p$ = 0.547) were not significant.

### 5.1.3 | Discussion

Study 2a confirms our initial theorizing such as the use of directive language by a brand can generate vicarious embarrassment. Although modest, this effect was positive and significant. Moreover, the findings showed no interaction effect of directiveness and conversation topic on vicarious embarrassment. As our initial theorizing was that conversation topic played a role in consumer response to directive band language, this left us with the possibility that while not moderating a X-M path, conversation topic could still moderate an X-Y path (with Y as engagement and M as vicarious embarrassment) as predicted. This relation should be tested in study 2b.

## 5.2 | Study 2b

### 5.2.1 | Method

The design of study 2b is a two (brand language: directive language use vs. control) × two (conversation topic: product-centered vs. nonproduct-centered) between-subjects experiment. Two hundred seventy-five participants ($M_{age}$ = 32.56, SD = 10.12; 49.8% male, 48.4% female, 1.5% other, and 0.4% prefer not to answer) were recruited from Prolific, and they received market rate financial compensation. The participants were all native French speakers from Western French-speaking countries (56.73% France, 25.45% Canada, 12.73% Belgium, 3.27% Switzerland, and 1.82% Luxemburg). They were screened as having declared that they were active Facebook users and were randomly assigned to the different experimental cells. We asked the participants to imagine that they were browsing their Facebook newsfeed; then, we exposed them to a stimulus, that is, a fictitious conversation between McDonald's brand and a consumer (Supporting Information S1: Web Appendix B).

The conversation topic was manipulated by creating a conversation about a new product from the brand (new vegetables offered in McDonald's restaurants; topic = product-centered) and a conversation about a broader topic (a text making a connection between sports and health; topic = nonproduct-centered). Brand use of directive language was manipulated by including directive language in the brand conversation (e.g.,

"Tell us how you'd like them cooked!"/"Tell us how long you run!") or by not including directive language (control).

Following the scenario, the participants answered the same questions as in study 2a to indicate the extent to which they perceived the presence of directive language ($\alpha = 0.86$; see Supporting Information S1: Web Appendix D), what the conversation topic was, and the extent to which they felt vicarious embarrassment ($\alpha = 0.95$). Moreover, as a proxy for consumer engagement (and, in essence, a holistic summary of the performed page interactions), an item assessed the extent to which respondents intended to interact with the brand after observing the conversation between McDonald's brand and a consumer (1 = "*never*"; 7 = "*all the time*"). This measure combines the behavioral dimension of consumer engagement (Dessart et al. 2016) with the single-item approach advocated by Bergkvist and Rossiter (2007).

### 5.2.2 | Results

We checked the manipulations by first conducting a one-way ANOVA. The findings revealed a significant effect of directive language use on directiveness perception ($F(1, 273) = 147.26$, $p < 0.001$), such that directive language perception was greater in the directive condition ($M_{directive} = 5.55$, SD = 1.09) than in the control condition ($M_{control} = 3.80$, SD = 1.29). Second, as in study 2a, we used the question about the conversation topic as manipulation and attention checks, as we screened participants on the basis of their attention by keeping only those who remembered the conversation topic correctly. Thus, as expected, the relationship between variables (i.e., manipulated topic and perceived topic) was significant ($\chi^2$ (1, N = 275) = 275, $p < 0.001$). Both results prove that the manipulations were successful. Finally, we checked the credibility of the stimuli using an ad hoc single-item question (Bergkvist and Rossiter 2007). The overall score was satisfactory ($M = 4.03$, SD = 1.77), and a one-way ANOVA found no significant differences between product-centered ($M_{product} = 4.07$, SD = 1.78) and nonproduct-centered conversations ($M_{nonproduct} = 3.99$, SD = 1.77; $F(1, 273) = 0.12$, $p = 0.725$).

To test the direct effect of directive language, we conducted a one-way MANOVA with directive language use as the independent variable, and consumer engagement and vicarious embarrassment as the dependent variables. The findings revealed a marginally significant and negative effect of directive language use on engagement ($F(1, 273) = 3.42$ $p = 0.065$, $\eta_p^2 = 0.012$), such as the use of directive language lowering engagement ($M_{directive} = 3.29$, SE = 0.11) compared with the control condition ($M_{control} = 3.57$, SE = 0.11). These results provide additional support for H1. Moreover, the findings revealed a significant and positive effect of directive language use on vicarious embarrassment ($F(1, 273) = 3.94$ $p = 0.048$, $\eta_p^2 = 0.014$), such as the use of directive language increasing vicarious embarrassment ($M_{directive} = 2.88$, SE = 0.14) compared with the control condition ($M_{control} = 2.49$, SE = 0.14).

To test the potential mediating effect of vicarious embarrassment and the moderating effect of the conversation topic, we conducted a bootstrapped moderation analysis via the PROCESS macro (Model 5; Hayes 2017) with 5000 bootstrap samples. Directive language use was the independent variable, with consumer engagement as the dependent variable, vicarious embarrassment as the mediating variable, and conversation topic as the moderating variable of the effect of directive language use on engagement.

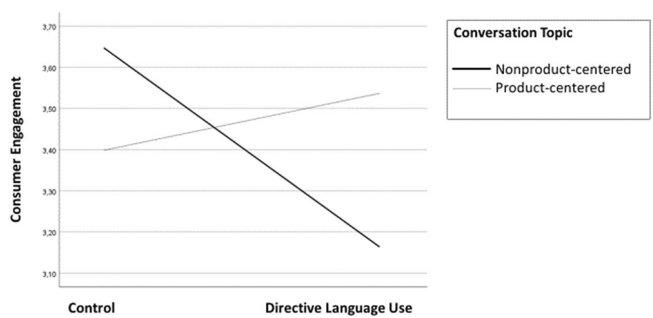

**FIGURE 2** | Interaction effect of directive language use and conversation topic on consumer engagement.

First, when the mediating vicarious embarrassment variable was included in the model, the results confirmed the significant and positive effect of directive language use on vicarious embarrassment ($b = 0.39$, $t = 1.99$, 95% CI [0.003, 0.772], $p = 0.048$). We also found that directive language use had a significant and negative indirect effect on engagement through vicarious embarrassment (X-M × M-Y = −0.10, 95% CI [−0.205, −0.003]); the confidence interval did not include zero, indicating a significant effect. These results provide support for H2.

Second, the results also revealed an interaction effect of directive language and conversation topic on consumer engagement ($b = -0.62$, $t = -2.12$, 95% CI [−1.20, −0.04], $p = 0.035$). Simple slope analysis indicated that when the conversation topic was product-centered, there was no significant effect of directive language use on consumer engagement ($b = 0.14$, $t = 0.65$, 95% CI [−0.28, 0.56], $p = 0.516$), whereas this relationship was significant and negative when the conversation topic was nonproduct-centered ($b = -0.48$, $t = -2.39$, 95% CI [−0.88, −0.08], $p = 0.018$, Figure 2). Indeed, when the conversation topic was nonproduct-centered, consumer engagement was lower ($M_{directive} = 3.16$, SE = 0.11) when the brand used directive language, compared with control ($M_{control} = 3.65$, SE = 0.14). These results support H3.

### 5.2.3 | Discussion

Study 2b explores the effect of directive language on consumer engagement through the mediating role of vicarious embarrassment and according to the topic of the conversation between a brand and a consumer. Facework theory predicts a negative impact of directive language, and the results confirm this negative impact on consumer engagement explained by the vicarious embarrassment felt by consumers with regard to the consumer who is addressed in a directive way. Moreover, the findings highlight the moderating role of conversation topic. Directive brand language use has a negative effect on consumer engagement when the conversation is nonproduct-centered (i.e., when the topic is the link between sports and health)

## 6 | Study 3: Moderating Role of Brand Relationship Strength

The purpose of study 3 is to test the moderating effect of brand relationship strength on the relationship between directive language use and consumer engagement (H4). The design is a 2 (brand language: directive language use vs. control) × 2 (brand relationship strength: high vs. low) between-subjects experiment. Since we predicted and found in study 2b that the negative effect of directive language occurs primarily in nonproduct-centered conversations, this experiment takes place in such a context.

### 6.1 | Method

One hundred and twenty-one participants ($M_{age} = 30.41$, SD = 8.75; 56.2% female, 43.8% male) were recruited from an online consumer panel and received market rate financial compensation. The participants all lived in France. They received an email with a link to a self-assessed survey where they were screened by confirming that they held an Instagram account. To manipulate brand relationship strength, we randomly asked participants to provide a brand they loved or a brand they did not love among five well-known brands in the food sector (Bonne Maman, Kinder, Lu, Milka, and Nestlé). The experimental platform incorporated the selected brand into pre-designed stimuli shown to participants. The participants were asked to answer questions about the brand, followed by a scenario in which they were asked to imagine browsing their Instagram newsfeed and noticing a particular brand–consumer conversation about a pancake recipe, which is not one of the brand products—that is, the conversation was nonproduct-centered. A realistic (though mock) Instagram conversation for the chosen brand was presented, featuring a brand-created post and a series of comments exchanged between the brand and a consumer. In the directive scenario, the brand was directly addressing a consumer within the comment section, asking and encouraging them to share their own content and to comment back (Supporting Information S1: Web Appendix C). Finally, the survey asked several questions about the conversation and about the participants.

As this study features brand love as a proxy for brand relationship strength, the participants indicated the extent to which they loved the brand they chose by answering six items with seven-point Likert scales (e.g., "*I feel emotionally connected to [Brand]*"; Bagozzi et al. 2017; $\alpha = 0.95$). After the participants were exposed to the scenario, we measured directiveness perception ($\alpha = 0.94$), as in study 2b. To measure consumer engagement, we used seven-point Likert scales combining three items from Schamari and Schaefers (2015; e.g., "I would participate in this discussion") with an item incorporating sharing intentions to the engagement construct as suggested by Obilo et al. (2021; "I would share this discussion to my Facebook account"; $\alpha = .95$). All measures are presented in the Supporting Information S1: Web Appendix D.

### 6.2 | Results

As a manipulation check, two-way MANOVA revealed that directive message use was perceived as expected ($F(1, 119) = 5.88$, $p = 0.017$), as was brand relationship strength ($F(1, 119) = 103.10$, $p < 0.001$). The participants in the directive condition reported a greater directiveness perception ($M_{directive} = 5.75$, SD = 1.26) than the participants in the control condition did ($M_{control} = 5.12$, SD = 1.58). In addition, participants in the high brand relationship strength condition reported more love for the brand ($M_{high\_relationships} = 5.63$, SD = 1.04) than did participants in the low brand relationship strength condition ($M_{low\_relationships} = 3.34$, SD = 1.50). These results prove that the manipulations were successful. Moreover, when answering a dichotomous scale asking them what was the conversation about (1 = [Brand]'s products, 7 = another topic), participants understood overall that the conversation was not about the brand's products ($M = 5.17$, SD = 1.96). Finally, we checked the credibility ($M = 5.46$, SD = 1.49) and realism ($M = 5.40$, SD = 1.48) of the stimuli using ad hoc single-item questions (Bergkvist and Rossiter 2007).

For hypothesis testing, we first conducted a two-way ANOVA with directive language use and brand relationship strength as independent variables and consumer engagement as the dependent variable. The findings revealed no significant main effect of directive language use ($p = 0.084$) and a significant main effect of brand relationship strength ($F(1, 117) = 6.95$, $p = 0.010$, $\eta_p^2 = 0.056$). The findings also revealed a significant interaction effect of directive language use and brand relationship strength ($F(1, 117) = 5.18$, $p = 0.025$, $\eta_p^2 = 0.042$; Figure 3). In the low relationship condition, participants exposed to directive language reported less intention to engage ($M_{directive} = 3.47$, SE = 0.32) than in the control condition ($M_{control} = 4.39$, SE = 0.36). In contrast, in the high relationship condition, participants exposed to directive language reported more intention to engage ($M_{directive} = 5.35$, SE = 0.30) than in the control condition ($M_{control} = 4.51$, SE = 0.33). These results support H4.

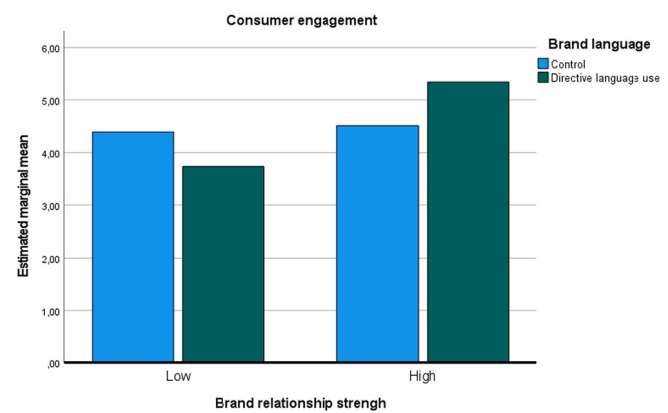

**FIGURE 3** | Interaction effect of directive language use and brand relationship strength on consumer engagement.

### 6.3 | Discussion

Study 3 revealed that, in a nonproduct-centered conversation, directive brand language decreases consumer engagement when consumers have weak relationships with the brand, while it increases engagement for those with strong brand relationships. Thus, brand relationship strength serves as a mitigator of the negative effect of directive brand language use.

### 7 | General Discussion

Our findings reveal the conditions under which a more or less positive effect of directive language occurs in brand conversations. As predicted by facework theory, we found that the use of directive language can have a detrimental effect on the engagement of consumers, who feel vicarious embarrassment when they observe a brand addressing other consumers in a directive way. The negative effect of directive language especially occurs when the conversation is nonproduct-centered. However, in this context, this negative effect is mitigated when consumers have a strong relationship with the brand.

### 7.1 | Theoretical Contributions

First, this study provides a valuable addition to the literature on directive language. While previous research has shown that recipients of directive messages need to escape pressure to comply (Fitzsimons and Lehmann 2004; Zemack-Rugar et al. 2017), the present research highlights that consumers expect brands to conform to conversation norms. Within this study, we consider directive language as FTAs, as opposed to Face-Flattering Acts, which have been studied previously (Fombelle et al. 2016). In this context, we show that consumers feel vicarious embarrassment when observing directive brand language directed at another consumer and can react negatively. Consumers may feel vicarious embarrassment when witnessing a violation of social norms in consumer–brand conversation. By generating an uncomfortable state, a brand conversation using directive language provokes less consumer engagement than does a conversation without the use of directive language. With such findings, this study also adds to the emerging streams of work on brand language, highlighting the importance of linguistic devices in branding efforts (Carnevale et al. 2017; Packard et al. 2024).

Second, our results contribute to the literature on brand–consumer interaction and engagement. Previous research on directive language has focused on the recipients of messages and on single-message strategies. Conversely, our research focuses on consumers who observe brands interacting with other consumers in entire brand conversations. While previous studies mention message type (De Vries et al. 2012; Irmak et al. 2020; Lee et al. 2018; Pezzuti et al. 2021), consumer psychological characteristics (Kavvouris et al. 2020), and consumer–brand relationships (Vafainia et al. 2019; Zemack-Rugar et al. 2017) as antecedents of consumer engagement, our research highlights a new variable: conversation topic. The present research shows that the inclusion of directive language in brand conversation has more or less positive effects on consumers, depending on whether the conversation is product-centered or nonproduct-centered.

Third, our findings contribute to facework theory by highlighting the conversation topic as an additional variable that guides interaction strategies beyond the relationship-, power-, and culture-related variables (Brown and Levinson 1987). In particular, in this context of product-centered conversation, consumers tend to tolerate directive language and show greater engagement toward the brand. A possible explanation is that when the conversation is about the core product, consumers categorize it as a form of advertising instead of a "real" conversation with face issues (Berthelot-Guiet 2020), making face loss unlikely.

### 7.2 | Managerial Implications

As directive language use is a popular technique among digital marketers in brand interactions (e.g., calls-to-action in online advertisements), it is important for practitioners to know when it is relevant to engage consumers and when it is not. On the one hand, our results suggest that, to drive more consumer engagement (i.e., likes, comments, shares) with social media conversations, marketers can use directive language within product-related discussions (e.g., new product launches). Indeed, by providing useful information and effective help to consumers, a direct style could encourage consumers to interact with the brand. On the other hand, when the conversation topic is not related to the core product, for example, when running "brand content" operations (Arrivé 2022), brands should give consumers a great deal of freedom by avoiding directive language.

Moreover, our results show that when brands interact with consumers who are engaged in strong brand relationships, directive language is tolerated. Thus, marketers can feel more comfortable using directive language when their brand conversation is seen by loyal consumers, whereas brands should be more cautious when their posts are seen by prospects. Such segmentation can be performed when a conversation is started by using sponsored posts targeted at a specific audience. For example, sponsored posts featuring directive language could be targeted at consumers who already "liked" the brand page. In addition, such language could be more appropriate in spaces dedicated to loyal consumers (e.g., private brand community groups) than in spaces that are open to a wider audience (e.g., public brand social media pages).

We also shed light on the need for adaptive strategies in digital communications. Indeed, brand spokespersons are supposed to behave consistently from one channel to another, as consumers use a range of different channels to address brands (Hamilton et al. 2016). However, our research shows that, within a single channel, brand representatives should adapt their interaction style to the context, for example, by modulating their communication style depending on the conversation topic and the relationships between brands and consumers. These findings are important for marketers who perform real-time marketing interventions on social media (Borah et al. 2020).

By demonstrating the importance of context as a condition for the use of directive brand language, this study provides managerial guidance for marketers, helping them to avoid creating a negative effect through the use of directive language in conversation. Our findings add to the body of research on brands' conversational faux pas. Indeed, when brands use language that is perceived as inauthentic, they risk negative consumer responses, including backlash, brand boycotts, and damaged credibility (Mirzaei et al. 2022; Fernando et al. 2014). This could be the case if brands are overly directive in a context where they are not expected to be.

Finally, this study highlights the potential effect of brand language in terms of consumer engagement. In particular, we found that brands that participate in nonproduct-centered conversation should avoid the use of directive language. This has important consequences for brands that are involved in brand activism by spreading cause-related conversations (e.g., about social, environmental, or health issues; Vredenburg et al. 2020). Encouraging consumers to engage is critical to such a strategy, as consumer comments can help increase activism, support brands, and educate other consumers (Fletcher-Brown et al. 2024). We provide evidence that brand language can influence both the valence ("likes" suggesting positive engagement) and the volume of engagement. We even observed the latter in a social issue-related context, as two studies featured a conversation about health (practicing sport). However, more research is needed to explore what types of user-generated content result from brands' use or avoidance of directive language.

## 7.3 | Limitations and Directions for Further Research

This study has several limitations. First, we focus on one dimension of consumer engagement—behavior—through the intention to participate in brand conversations and the intention to share discussions on personal social media accounts. However, directive language in brand conversation can also impact the emotional and cognitive dimensions of engagement. Cognitive engagement refers to the degree of psychological effort invested in an interaction (attention, concentration, and absorption), whereas emotional engagement refers to the degree of affective attachment dedicated to an object (Dessart et al. 2016; Elmashhara et al. 2024), such as a robot or a conversation. On this basis, future research can examine the influence of directive language on the attention/absorption dedicated to brand conversations and explore how it plays a role in the enthusiasm/enjoyment dedicated to online brand conversations.

Second, our research focuses on the impact of directive language in conversations between a brand and a consumer. However, the influence of content depends on the nature of its source (Lou et al. 2019). For example, a brand may be perceived as less human than a social media influencer (Kim and Kim 2022). Future research could investigate the use of directive language by influencers in social media, where relational conversations are common and where consumers can feel a sense of belonging to a community. In this context, consumers may believe that the recipients of directive brand messages partially accept brand influence through compliance (Hollebeek et al. 2022) and then may feel less vicarious embarrassment.

Thus, future research could investigate the use of directive language by influencers on consumers who feel that they belong to the same community as the recipients of directive messages.

In another vein, we focused on the food market and on business-to-consumer brands. Therefore, future research on directive language on social media could consider other business areas and explore business-to-business contexts. Finally, our focus is the response from consumers observing public interactions. Thus, future research could study the response of actual recipients of directive brand language in both public and private settings.

In conclusion, this study enriches the knowledge of how consumers perceive directive language in online conversations with brands. As social media is still expected to have an important impact on brand activities in the future (Appel et al. 2020), research on consumer responses to social media marketing efforts still seems promising.

**Acknowledgments**

The authors would like to thank Orline Poulat, research engineer at the Maison des Sciences de l'Homme Lyon-Saint-Etienne, for her help in collecting and analyzing field data, and the Chair Brands, Values & Society at IAE Paris-Sorbonne for its support. The authors also thank the marketing faculty at Columbia Business School for their valuable advice during the first author's visit, and especially Donald Lehmann for his thoughtful comments on earlier drafts.

**Data Availability Statement**

The data that support the findings of this study are available from the corresponding author upon reasonable request.

## Endnotes

[1] To conduct the sentiment analysis of our posts, we used Levallois (2013) nocodefunctions/Umigon tool and API. This tool employs a rule-based system to analyze text and determine whether the sentiment is neutral, positive, or negative. According to a comparative analysis of 23 alternatives, nocodefunctions/Umigon was found to be particularly effective for analyzing social media text, such as posts on X (Ribeiro et al. 2016).

[2] "Retweets" are now called "reposts" since Twitter's rebranding to X; a "quote" is a specific retweet that allows users to add a comment while sharing a tweet.

[3] Demographics such as age and gender were collected and employed as covariates in our analyses across studies. We also checked for the participant's tendency to engage with social media content on Facebook (study 2a) and Instagram (study 3). None had an impact on the observed results and will therefore not be mentioned further.

## References

Aggarwal, P. 2004. "The Effects of Brand Relationship Norms on Consumer Attitudes and Behavior." *Journal of Consumer Research* 31, no. 1: 87–101.

Aguinis, H., and K. J. Bradley. 2014. "Best Practice Recommendations for Designing and Implementing Experimental Vignette Methodology Studies." *Organizational Research Methods* 17, no. 4: 351–371.

Allwood, J. 1977. "A Critical Look at Speech Act Theory." In *Logic, Pragmatics and Grammar*, edited by Ö. Dahl, 53–99. Studentlitteratur.

Andriuzzi, A., and G. Michel. 2021. "Brand Conversation: Linguistic Practices on Social Media in the Light of Face-Work

Theory." *Recherche et Applications en Marketing (English Edition)* 36, no. 1: 44–64.

Appel, G., L. Grewal, R. Hadi, and A. T. Stephen. 2020. "The Future of Social Media in Marketing." *Journal of the Academy of Marketing Science* 48, no. 1: 79–95.

Arrivé, S. 2022. "Digital Brand Content: Underlying Nature and Rationales of a Hybrid Marketing Practice." *Journal of Strategic Marketing* 30, no. 4: 340–354.

Ashley, C., and T. Tuten. 2015. "Creative Strategies in Social Media Marketing: An Exploratory Study of Branded Social Content and Consumer Engagement." *Psychology & Marketing* 32, no. 1: 15–27.

Austin, J. L. 1962. *How to Do Things With Words*. Oxford University Press.

Bagozzi, R. P., R. Batra, and A. Ahuvia. 2017. "Brand Love: Development and Validation of a Practical Scale." *Marketing Letters* 28, no. 1: 1–14.

Baldus, B. J., C. Voorhees, and R. Calantone. 2015. "Online Brand Community Engagement: Scale Development and Validation." *Journal of Business Research* 68, no. 5: 978–985.

Batra, R., A. Ahuvia, and R. P. Bagozzi. 2012. "Brand Love." *Journal of Marketing* 76, no. 2: 1–16.

Ben, Z., and P. Shukla. 2024. "They Forgot Me! The Exclusionary Effects Among Complaining Consumers When Others Receive a Response." *Psychology & Marketing* 41, no. 11: 2741–2756.

Berger, J., A. Humphreys, S. Ludwig, W. W. Moe, O. Netzer, and D. A. Schweidel. 2020. "Uniting the Tribes: Using Text for Marketing Insight." *Journal of Marketing* 84, no. 1: 1–25.

Bergkvist, L., and J. R. Rossiter. 2007. "The Predictive Validity of Multiple-Item Versus Single-Item Measures of the Same Constructs." *Journal of Marketing Research* 44, no. 2: 175–184.

Berthelot-Guiet, K. 2020. "The Digital 'Advertising Call': An Archeology of Advertising Literacy." In *Social Computing and Social Media. Participation, User Experience, Consumer Experience, and Applications of Social Computing*, edited by G. Meiselwitz, Lecture Notes in Computer Science, vol. 12195, 278–294. Springer.

Borah, A., S. Banerjee, Y. T. Lin, A. Jain, and A. B. Eisingerich. 2020. "Improvised Marketing Interventions in Social Media." *Journal of Marketing* 84, no. 2: 69–91.

Brehm, J. W. 1966. *A Theory of Psychological Reactance*. Academic Press.

Brown, P., and S. C. Levinson. 1987. *Politeness: Some Universals in Language Use*. Cambridge University Press.

Campbell, C., C. Ferraro, and S. Sands. 2014. "Segmenting Consumer Reactions to Social Network Marketing." *European Journal of Marketing* 48, no. 3/4: 432–452.

Carnevale, M., D. Luna, and D. Lerman. 2017. "Brand Linguistics: A Theory-Driven Framework for the Study of Language in Branding." *International Journal of Research in Marketing* 34, no. 2: 572–591.

Crowdfire. 2022. "The Power of the Call-to-Action in Social Media, by Oscar." https://read.crowdfireapp.com/2022/06/10/the-power-of-the-call-to-action-in-social-media/.

Dahl, D. W., R. V. Manchanda, and J. J. Argo. 2001. "Embarrassment in Consumer Purchase: The Roles of Social Presence and Purchase Familiarity." *Journal of Consumer Research* 28, no. 3: 473–481.

Dessart, L., C. Veloutsou, and A. Morgan-Thomas. 2015. "Consumer Engagement in Online Brand Communities: A Social Media Perspective." *Journal of Product & Brand Management* 24, no. 1: 28–42.

Dessart, L., C. Veloutsou, and A. Morgan-Thomas. 2016. "Capturing Consumer Engagement: Duality, Dimensionality and Measurement." *Journal of Marketing Management* 32, no. 5–6: 399–426.

Doan, A. 2023. "Best Brands on Social Media: 10 Inspiring Examples to Follow." Nextiva. https://www.nextiva.com/blog/best-brands-on-social-media.html.

Donovan, L., J. Priester, D. MacInnis, and W. Park. 2012. "Brand Forgiveness: How Close Brand Relationships Influence Forgiveness." In *Consumer-Brand Relationships: Theory and Practice*, edited by S. Fournier, M. Breazeale and M. Fetscherin, 184–203. Routledge.

Ellison, N. B., P. Triệu, S. Schoenebeck, R. Brewer, and A. Israni. 2020. "Why We Don't Click: Interrogating the Relationship Between Viewing and Clicking in Social Media Contexts by Exploring the 'Non-Click'." *Journal of Computer-Mediated Communication* 25, no. 6: 402–426.

Elmashhara, M. G., R. De Cicco, S. C. Silva, M. Hammerschmidt, and M. L. Silva. 2024. "How Gamifying AI Shapes Customer Motivation, Engagement, and Purchase Behavior." *Psychology & Marketing* 41, no. 1: 134–150.

Fernando, A. G., L. Suganthi, and B. Sivakumaran. 2014. "If You Blog, Will They Follow? Using Online Media to Set the Agenda for Consumer Concerns on "Greenwashed" Environmental Claims." *Journal of Advertising* 43, no. 2: 167–180.

Fitzsimons, G. J., and D. R. Lehmann. 2004. "Reactance to Recommendations: When Unsolicited Advice Yields Contrary Responses." *Marketing Science* 23, no. 1: 82–94.

Fletcher-Brown, J., K. Middleton, H. Thompson-Whiteside, S. Turnbull, A. Tuan, and L. D. Hollebeek. 2024. "The Role of Consumer Speech Acts in Brand Activism: A Transformative Advertising Perspective." *Journal of Advertising* 53, no. 4: 491–510.

Fombelle, P. W., S. A. Bone, and K. N. Lemon. 2016. "Responding to the 98%: Face-Enhancing Strategies for Dealing With Rejected Customer Ideas." *Journal of the Academy of Marketing Science* 44: 685–706.

Fournier, S. 1998. "Consumers and Their Brands: Developing Relationship Theory in Consumer Research." *Journal of Consumer Research* 24, no. 4: 343–353.

Gavilanes, J. M., T. C. Flatten, and M. Brettel. 2018. "Content Strategies for Digital Consumer Engagement in Social Networks: Why Advertising Is an Antecedent of Engagement." *Journal of Advertising* 47, no. 1: 4–23.

Goffman, E. 1955. "On Facework: An Analysis of Ritual Elements in Social Interaction." *Psychiatry* 18, no. 3: 213–231.

Goffman, E. 1956. "Embarrassment and Social Organization." *American Journal of Sociology* 62, no. 3: 264–271.

Goffman, E. 1967. *Interaction Ritual: Essays on Face-to-face Interaction*. Anchor Books.

Hamilton, M., V. D. Kaltcheva, and A. J. Rohm. 2016. "Hashtags and Handshakes: Consumer Motives and Platform Use in Brand-Consumer Interactions." *Journal of Consumer Marketing* 33, no. 2: 135–144.

Hamilton, R. W., A. Schlosser, and Y.-J. Chen. 2017. "Who's Driving This Conversation? Systematic Biases in the Content of Online Consumer Discussions." *Journal of Marketing Research* 54, no. 4: 540–555.

Hayes, A. F. 2017. *Introduction to Mediation, Moderation, and Conditional Process Analysis: A Regression-based Approach*. Guilford Press.

Hollebeek, L. D., M. S. Glynn, and R. J. Brodie. 2014. "Consumer Brand Engagement in Social Media: Conceptualization, Scale Development and Validation." *Journal of Interactive Marketing* 28, no. 2: 149–165.

Hollebeek, L. D., D. E. Sprott, V. Sigurdsson, and M. K. Clark. 2022. "Social Influence and Stakeholder Engagement Behavior Conformity, Compliance, and Reactance." *Psychology & Marketing* 39, no. 1: 90–100.

Homburg, C., L. Ehm, and M. Artz. 2015. "Measuring and Managing Consumer Sentiment in an Online Community Environment." *Journal of Marketing Research* 52, no. 5: 629–641.

Hootsuite. 2022. "How to Write a Great Social Media Call to Action, by Liz Stanton." https://blog.hootsuite.com/how-to-write-effective-ctas/.

Hubspot. 2024. "49 Call-to-Action Examples You Can't Help but Click, by Brittany Leaning." https://blog.hubspot.com/marketing/call-to-action-examples.

Hudson, S., L. Huang, M. S. Roth, and T. J. Madden. 2016. "The Influence of Social Media Interactions on Consumer–Brand Relationships: A Three-Country Study of Brand Perceptions and Marketing Behaviors." *International Journal of Research in Marketing* 33, no. 1: 27–41.

Irmak, C., M. R. Murdock, and V. K. Kanuri. 2020. "When Consumption Regulations Backfire: The Role of Political Ideology." *Journal of Marketing Research* 57, no. 5: 966–984.

Jakic, A., M. O. Wagner, and A. Meyer. 2017. "The Impact of Language Style Accommodation During Social Media Interactions on Brand Trust." *Journal of Service Management* 28, no. 3: 418–441.

Kaplan, A. M., and M. Haenlein. 2010. "Users of the World, Unite! The Challenges and Opportunities of Social Media." *Business Horizons* 53, no. 1: 59–68.

Kavvouris, C., P. Chrysochou, and J. Thøgersen. 2020. "'Be Careful What You Say': The Role of Psychological Reactance on the Impact of Pro-Environmental Normative Appeals." *Journal of Business Research* 113: 257–265.

Keller, K. L. 1993. "Conceptualizing, Measuring, and Managing Customer-Based Brand Equity." *Journal of Marketing* 57, no. 1: 1–22.

Kilian, T., S. Steinmann, and E. Hammes. 2018. "Oh My Gosh, I Got to Get out of This Place! A Qualitative Study of Vicarious Embarrassment in Service Encounters." *Psychology & Marketing* 35, no. 1: 79–95.

Kim, B., and O. Velthuis. 2021. "From Reactivity to Reputation Management: Online Consumer Review Systems in the Restaurant Industry." *Journal of Cultural Economy* 14, no. 6: 675–693.

Kim, D. Y., and H. Y. Kim. 2022. "Social Media Influencers as Human Brands: An Interactive Marketing Perspective." *Journal of Research in Interactive Marketing* 17, no. 1: 94–109.

Kim, S. J., R. J.-H. Wang, E. Maslowska, and E. C. Malthouse. 2016. "Understanding a Fury in Your Words": The Effects of Posting and Viewing Electronic Negative Word-Of-Mouth on Purchase Behaviors." *Computers in Human Behavior* 54: 511–521.

Köhler, C. F., A. J. Rohm, K. De Ruyter, and M. Wetzels. 2011. "Return on Interactivity: The Impact of Online Agents on Newcomer Adjustment." *Journal of Marketing* 75, no. 2: 93–108.

Kronrod, A., A. Grinstein, and L. Wathieu. 2012. "Go Green! Should Environmental Messages Be So Assertive?" *Journal of Marketing* 76, no. 1: 95–102.

Kumar, A., R. Bezawada, R. Rishika, R. Janakiraman, and P. K. Kannan. 2016. "From Social to Sale: The Effects of Firm-Generated Content in Social Media on Customer Behavior." *Journal of Marketing* 80, no. 1: 7–25.

Lee, D., K. Hosanagar, and H. S. Nair. 2018. "Advertising Content and Consumer Engagement on Social Media: Evidence From Facebook." *Management Science* 64, no. 11: 5105–5131.

Levallois, C. 2013. "Umigon: Sentiment Analysis for Tweets Based on Terms Lists and Heuristics." In *Proceedings of the Seventh International Workshop on Semantic Evaluation (SemEval)*, edited by S. Manandhar and D. Yuret, 414–417. Association for Computational Linguistics.

Lou, C., S. S. Tan, and X. Chen. 2019. "Investigating Consumer Engagement With Influencer-Vs. Brand-Promoted Ads: The Roles of Source and Disclosure." *Journal of Interactive Advertising* 19, no. 3: 169–186.

Marketing Charts. 2022. "Leading Ways to Engage With Brands on Social Media in Exchange for an Incentive According to Consumers in the United States as of June 2022." Statista. https://www.statista.com/statistics/1373044/top-ways-engage-brands-exchange-incentive-us/.

Meire, M., K. Coussement, A. De Caigny, and S. Hoornaert. 2022. "Does It Pay Off to Communicate Like Your Online Community? Evaluating the Effect of Content and Linguistic Style Similarity on B2B Brand Engagement." *Industrial Marketing Management* 106: 292–307.

Miller, R. S. 1987. "Empathic Embarrassment: Situational and Personal Determinants of Reactions to the Embarrassment of Another." *Journal of Personality and Social Psychology* 53, no. 6: 1061–1069.

Mirzaei, A., D. C. Wilkie, and H. Siuki. 2022. "Woke Brand Activism Authenticity or the Lack of It." *Journal of Business Research* 139: 1–12.

Mousavi, S., and S. Roper. 2023. "Enhancing Relationships Through Online Brand Communities: Comparing Posters and Lurkers." *International Journal of Electronic Commerce* 27, no. 1: 66–99.

Van Noort, G., and L. M. Willemsen. 2012. "Online Damage Control: The Effects of Proactive Versus Reactive Webcare Interventions in Consumer-Generated and Brand-Generated Platforms." *Journal of Interactive Marketing* 26, no. 3: 131–140.

Obilo, O. O., E. Chefor, and A. Saleh. 2021. "Revisiting the Consumer Brand Engagement Concept." *Journal of Business Research* 126: 634–643.

Oltra, I., C. Camarero, and R. San José Cabezudo. 2022. "Inspire Me, Please! The Effect of Calls to Action and Visual Executions on Customer Inspiration in Instagram Communications." *International Journal of Advertising* 41, no. 7: 1209–1234.

Packard, G., Y. Li, and J. Berger. 2024. "When language matters." *Journal of Consumer Research* 51, no. 3: 634–653.

Park, J. 2008. "Linguistic Politeness and Face-Work in Computer Mediated Communication, Part 2: An Application of the Theoretical Framework." *Journal of the American Society for Information Science and Technology* 59, no. 14: 2199–2209.

Pezzuti, T., J. M. Leonhardt, and C. Warren. 2021. "Certainty in Language Increases Consumer Engagement on Social Media." *Journal of Interactive Marketing* 53, no. 1: 32–46.

Pogacar, R., L. J. Shrum, and T. M. Lowrey. 2018. "The Effects of Linguistic Devices on Consumer Information Processing and Persuasion: A Language Complexity × Processing Mode Framework." *Journal of Consumer Psychology* 28, no. 4: 689–711.

Qi, P., Y. Zhang, Y. Zhang, J. Bolton, and C. D. Manning. 2020. "Stanza: A Python Natural Language Processing Toolkit for Many Human Languages." Preprint, arXiv, arXiv:2003.07082.

Rahman, R., T. Langner, and D. Temme. 2021. "Brand Love: Conceptual and Empirical Investigation of a Holistic Causal Model." *Journal of Brand Management* 28, no. 6: 609–642.

Razzaq, A., W. Shao, and S. Quach. 2023. "Towards An Understanding of Meme Marketing: Conceptualisation and Empirical Evidence." *Journal of Marketing Management* 39, no. 7–8: 670–701.

Ribeiro, F. N., M. Araújo, P. Gonçalves, M. André Gonçalves, and F. Benevenuto. 2016. "Sentibench—A Benchmark Comparison of State-of-the-Practice Sentiment Analysis Methods." *EPJ Data Science* 5: 23.

Sangwan, V., M. Maity, S. Tripathi, and A. Chakraborty. 2024. "From Discomfort to Desirable: The Effect of Embarrassment on Prosocial Consumption." *Psychology & Marketing* 41, no. 8: 1820–1832.

Schamari, J., and T. Schaefers. 2015. "Leaving the Home Turf: How Brands Can Use Webcare on Consumer-Generated Platforms to Increase Positive Consumer Engagement." *Journal of Interactive Marketing* 30, no. 1: 20–33.

Searle, J. R. 1969. *Speech Acts: An Essay in the Philosophy of Language*. Cambridge University Press.

Searle, J. R. 1979. *Expression and Meaning: Studies in the Theory of Speech Acts*. Cambridge University Press.

Semrush. 2022. "Main Call-to-Action (CTA) on Amazon.com Ads Descriptions in the United States as of 2022, by Number of Occurrences." Statista. https://www.statista.com/statistics/1338096/ads-title-call-to-action-on-amazon/.

Septianto, F., and N. Garg. 2021. "The Impact of Gratitude (vs Pride) on the Effectiveness of Cause-Related Marketing." *European Journal of Marketing* 55, no. 6: 1594–1623.

Šerić, M., and D. G. Praničević. 2018. "Consumer-Generated Reviews on Social Media and Brand Relationship Outcomes in the Fast-Food Chain Industry." *Journal of Hospitality Marketing & Management* 27, no. 2: 218–238.

Smith, A. N., E. Fischer, and C. Yongjian. 2012. "How Does Brand-Related User-Generated Content Differ Across Youtube, Facebook, and Twitter?" *Journal of Interactive Marketing* 26, no. 2: 102–113.

Sprout Social. 2021. "Ways Consumers Engaged With Brands on Social Media in the United States as of April 2021." Statista. https://www.statista.com/statistics/1273890/consumer-engage-brands-social-media-usa/.

Sprout Social. 2023. "New Research Indicates a Shift in What Consumers Find Memorable on Social Media." Sprout Social. https://sproutsocial.com/insights/press/new-research-indicates-a-shift-in-what-consumers-find-memorable-on-social-media/.

Swani, K., and L. I. Labrecque. 2020. "Like, Comment, or Share? Self-Presentation Vs. Brand Relationships as Drivers of Social Media Engagement Choices." *Marketing Letters* 31, no. 2: 279–298.

Tan, T. M., M. S. Balaji, E. L. Oikarinen, S. Alatalo, and J. Salo. 2021. "Recover From a Service Failure: The Differential Effects of Brand Betrayal and Brand Disappointment on An Exclusive Brand Offering." *Journal of Business Research* 123: 126–139.

Vafainia, S., E. Breugelmans, and T. Bijmolt. 2019. "Calling Customers to Take Action: The Impact of Incentive and Customer Characteristics on Direct Mailing Effectiveness." *Journal of Interactive Marketing* 45, no. 1: 62–80.

Villanova, D., and T. Matherly. 2024. "For Shame! Socially Unacceptable Brand Mentions on Social Media Motivate Consumer Disengagement." *Journal of Marketing* 88, no. 2: 61–78.

Villarroel Ordenes, F., D. Grewal, S. Ludwig, K. D. Ruyter, D. Mahr, and M. Wetzels. 2019. "Cutting Through Content Clutter: How Speech and Image Acts Drive Consumer Sharing of Social Media Brand Messages." *Journal of Consumer Research* 45, no. 5: 988–1012.

Vredenburg, J., S. Kapitan, A. Spry, and J. A. Kemper. 2020. "Brands Taking a Stand: Authentic Brand Activism or Woke Washing?" *Journal of Public Policy & Marketing* 39, no. 4: 444–460.

De Vries, L., S. Gensler, and P. S. H. Leeflang. 2012. "Popularity of Brand Posts on Brand Fan Pages: An Investigation of the Effects of Social Media Marketing." *Journal of interactive marketing* 26, no. 2: 83–91.

Wang, C. X., and J. Zhang. 2020. "Assertive Ads for Want or Should? It Depends on Consumers' Power." *Journal of Consumer Psychology* 30, no. 3: 466–485.

Wang, Y., and A. Chaudhry. 2018. "When and How Managers' Responses to Online Reviews Affect Subsequent Reviews." *Journal of Marketing Research* 55, no. 2: 163–177.

Yule, G. 1996. *Pragmatics*. Oxford University Press.

Zemack-Rugar, Y., S. G. Moore, and G. J. Fitzsimons. 2017. "Just Do It! Why Committed Consumers React Negatively to Assertive Ads." *Journal of Consumer Psychology* 27, no. 3: 287–301.

Zhao, X., and Y. R. R. Chen. 2022. "How Brand-Stakeholder Dialogue Drives Brand-Hosted Community Engagement on Social Media: A Mixed-Methods Approach." *Computers in Human Behavior* 131: 107208.

Ziegler, A. H., A. M. Allen, J. Peloza, and J. Ian Norris. 2022. "The Nature of Vicarious Embarrassment." *Journal of Business Research* 153: 355–364.

**Supporting Information**

Additional supporting information can be found online in the Supporting Information section.

**WEB Appendix A:** Stimuli Used in Study 2a. **WEB Appendix B:** Stimuli Used in Study 2b. **WEB Appendix C:** Stimuli Used in Study 3. **WEB Appendix D:** Measures of Variables: Studies 2a, 2b, 3.